\documentclass[runningheads]{llncs}
\usepackage[ruled,vlined]{algorithm2e}
\usepackage{amsfonts}       
\usepackage{mathrsfs}
\usepackage{amsmath,amssymb}
\usepackage{graphicx}
\usepackage{tikz}
\usepackage{hyperref}
\usepackage{booktabs}
\usepackage{multirow}
\hypersetup{
    colorlinks=true,
    linkcolor=blue,
    filecolor=magenta,      
    urlcolor=cyan,
}
\newcommand{\beginsupplement}{%
        \setcounter{table}{0}
        \renewcommand{\thetable}{S\arabic{table}}%
        \setcounter{figure}{0}
        \renewcommand{\thesection}{\Alph{section}}
     }

\begin{document}

\title{Few-Shot Anomaly Detection for Polyp Frames from Colonoscopy }


%
%
\author{Yu Tian\inst{1,3}$\quad$
Gabriel Maicas\inst{1} $\quad$
Leonardo Zorron Cheng Tao Pu\inst{2,4}$\quad$
Rajvinder Singh\inst{2}$\quad$
Johan W. Verjans\inst{2,3}$\quad$
Gustavo Carneiro\inst{1}}


%
\authorrunning{Authors Suppressed Due to Excessive Length}
%
\institute{Australian Institute for Machine Learning, The University of Adelaide \and
Faculty of Health and Medical Sciences, The University of Adelaide  \and
South Australian Health and Medical Research Institute
\and Department of Gastroenterology and Hepatology, Nagoya University
}


\maketitle              
\begin{abstract}
\footnote{This work was partially supported by Australian Research Council grant DP180103232.}
Anomaly detection methods generally target the learning of a normal image distribution (i.e., inliers showing healthy cases) and during testing, samples relatively far from the learned distribution are classified as anomalies (i.e., outliers showing disease cases).
These approaches tend to be sensitive to outliers that lie relatively close to inliers (e.g., a colonoscopy image with a small polyp). 
In this paper, we address the inappropriate sensitivity to outliers by also learning from inliers. We propose a new few-shot anomaly detection method based on an encoder trained to maximise the mutual information between feature embeddings and normal images, followed by a few-shot score inference network, trained with a large set of inliers and a substantially smaller set of outliers.
We evaluate our proposed method on the clinical problem of detecting frames containing polyps from colonoscopy video sequences, where the training set has 13350 normal images (i.e., without polyps) and less than 100 abnormal images (i.e., with polyps). 
The results of our proposed model on this data set reveal a state-of-the-art detection result, while the performance based on different number of anomaly samples is relatively stable after approximately 40 abnormal training images. Code is available at \url{https://github.com/tianyu0207/FSAD-Net }. 
\end{abstract}

\section{Introduction}

\begin{figure}[t!]
\small
\begin{center}
 \includegraphics[width = \linewidth]{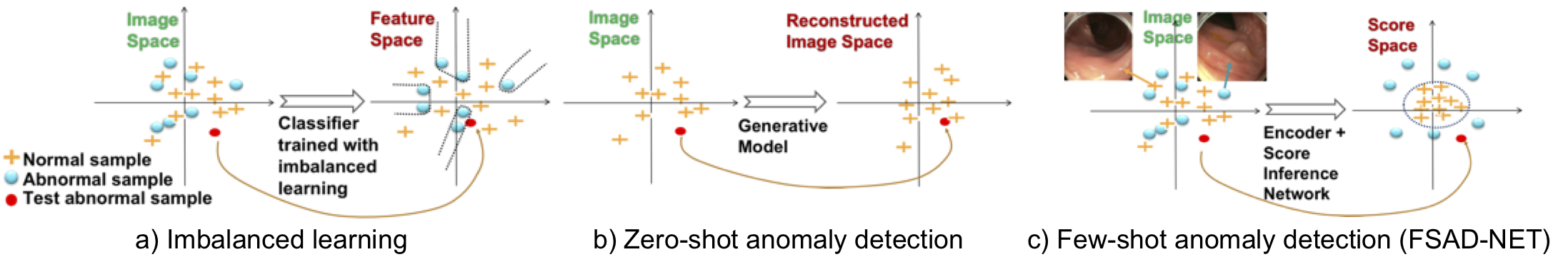}
\end{center}
\caption{Depiction of the three different approaches to handle few-shot and zero-shot anomaly detection. Our proposed FSAD-NET demonstrate better deviations between normal and abnormal samples}
\label{fig:intro_img}
\vspace{-10pt}
\end{figure}

Classification of rare events is a common problem in medical image analysis~\cite{litjens2017survey}, e.g., disease detection in medical screening tests such as colonoscopy.  
In this scenario, normal images generally come from healthy patients, while abnormal images are from unhealthy ones, where the proportion of normal images in the training set tends to be substantially larger than the abnormal ones.
One possible way to address such problems is through the design of training methods that can deal with imbalanced learning problems~\cite{li2019overfitting,lin2018focal} (Fig.~\ref{fig:intro_img}-(a)).  Even though they are often effective, these approaches still need a fairly high number of  abnormal training images.
Alternatively, zero-shot anomaly detection methods~\cite{gong2019memorizing,makhzani2015adversarial,zong2018deep,liu2019photoshopping} tackle this problem using a training set containing only normal images to train a conditional generative model that can reconstruct normal images, and anomalies are detected based on the reconstruction errors of testing images (Fig.~\ref{fig:intro_img}-(b)).  
Unfortunately, in practice these methods can misclassify outliers that lie relatively close to inliers (e.g., when cancer tissue occupies a small area of the image). Therefore, we propose a middle ground between these two approaches to address the issues of requiring a relatively large annotated data set and misclassifying challenging outliers. 

In this paper, we propose a few-shot anomaly detection method network (FSAD-NET) that is trained with a highly imbalanced training set, containing a large number of normal images (more than $10,000$) and few abnormal images (less than $100$) -- Fig.~\ref{fig:intro_img}-(c).  The method first learns a feature encoder that is trained with normal images to maximise the mutual information (MI) between the training images and feature embeddings~\cite{hjelm2018learning}.  Next, we train a score inference network (SIN)~\cite{pang2019deep} that pulls the feature embeddings of normal images close together toward a particular region of the feature space and pushes the embeddings of abnormal images away from that region of normal features.

In practice, FSAD-NET needs significantly less abnormal training images than typical imbalanced learning problems~\cite{li2019overfitting,lin2018focal}.  Moreover, given that we access a few abnormal training images, FSAD-NET has the potential to be more effective at correctly classifying challenging outliers compared to typical zero-shot anomaly detection methods~\cite{gong2019memorizing,makhzani2015adversarial,zong2018deep,liu2019photoshopping}.
To the best of our knowledge, our method is the first medical image analysis work to explore few-shot anomaly detection with a feature encoder that maximises MI between training images and embeddings, and explicitly optimises anomaly scores.
We evaluate FSAD-NET on the detection of colonoscopy video frames that contain polyps with a training set of more than 10000 normal images (without polyps) and less than 100 abnormal images. 
Results show that our FSAD-NET is more accurate than previous zero-shot anomaly detection approaches, which allows us to conclude that incorporating few abnormal cases into the training process improves the performance of anomaly detection methods.
Our approach also shows better accuracy than imbalanced learning methods, suggesting that FSAD-NET is more effective at dealing with very small training sets of abnormal images.
We will make our code publicly available (upon acceptance of our paper) to foster reproducibility and research on the area.

\section{Related Work}

Colorectal cancer is considered to be one of the most harmful cancers~\cite{siegel2014colorectal,tian2019one}. 
One effective method for screening patients for colorectal cancer is colonoscopy, where the goal is to detect polyps that are malignant or pre-malignant using a camera that is inserted into the bowel.
Accurate early detection of polyps may improve the 5-year survival rate to over 90\%~\cite{siegel2014colorectal}.  Unfortunately, the accuracy and speed of manual polyp detection can be affected by human factors, such as fatigue and expertise~\cite{pu2018sa1908,van2006polyp}. Therefore, automated polyp detection systems could help doctors improve polyp detection accuracy during a colonoscopy~\cite{yu_isbi19}.
Traditional systems to detect polyps are based on a supervised two-class classifier~\cite{korbar2017deep,yu_isbi19} trained with large training sets of images without polyps (i.e. normal) and images containing polyps (i.e. abnormal). 
Annotation of such training sets is unfortunately difficult because the vast majority of colonoscopy video frames contain normal images, making the manual search for images that contain polyps challenging.
Imbalanced learning solutions can therefore be used in this context~\cite{li2019overfitting,lin2018focal}, but its extension to polyp detection may not be effective without a relatively large number of abnormal images in the training set. 
Because of this limitation, zero-shot anomaly detection methods have been studied~\cite{schlegl2019f,schlegl2017unsupervised,perera2019ocgan,liu2018future,masci2011stacked,pang2019deep,liu2019photoshopping}, where the idea is to learn a distribution of normal images in a particular feature space, to subsequently test samples that do not fit well in this distribution and are then classified as an outlier that may contain a polyp.

Zero-shot anomaly detection methods assume that the conditional generative model~\cite{gong2019memorizing,makhzani2015adversarial,zong2018deep,schlegl2019f,schlegl2017unsupervised,perera2019ocgan,liu2019photoshopping}) can only reconstruct normal data.  Hence, when presented with an abnormal test image, the model produces a large reconstruction error. 
However, using an image reconstruction error for training is an indirect optimisation of the anomaly score, which can lead to a sub-optimal training process. For example, an abnormal image with a small polyp may have a low reconstruction error because the small area affected by the polyp and can be wrongly classified as normal.
We advocate that the performance of zero-shot anomaly detection methods can improve with the use of a small set of abnormal training images (less than $100$). Such imbalance learning problem has been tackled by few-shot classification approaches before. However, our problem has a different setup compared to problems handled by traditional few-shot learning methods that generally have many few-shot balanced multi-class problems for training~\cite{sung2018learning,nichol2018first,finn2017model}, while ours has only one few-shot highly imbalanced binary problem for training. Hence, we can only compare our method with  baseline approaches that handle imbalance learning~\cite{ren2018learning,lin2018focal}. For instance, Ren et al.~\cite{ren2018learning} propose a learning algorithm for highly imbalanced learning problems that weights training samples using a balanced validation set -- the need for this validation set is a disadvantage of this approach.  The focal loss approach~\cite{lin2018focal} is effective at handling imbalanced learning, but it may still need a large number of samples from both classes.

Few-shot anomaly detection has been shown in a non-medical image analysis context with the method SIN~\cite{pang2018learning} that is designed to directly optimise an anomaly score for normal and abnormal images. 
The main challenge to train SIN lies in the high dimensionality of the images~\cite{pang2018learning}.
Therefore, one way to alleviate this challenge is to introduce a dimensionality reduction before training SIN.  Recently, deep infomax (DIM)~\cite{hjelm2018learning} has been shown to be an effective dimensionality reduction approach. 
In our paper, we propose a method that uses DIM to learn a low-dimensionality feature embedding that is then used by SIN to classify anomalies. 

\section{Data Set and Method}


\subsection{Data set}
\label{sec:method_data}

The data set is obtained from 18 colonoscopy videos from 15 patients.  Video frames containing blurred visual information are removed using the variance of Laplacian method~\cite{he2006laplacian}. We then sub-sample consecutive frames by taking one of every five frames because the correlation between them makes training ineffective.  We also remove frames containing feces and water to reduce the need for a very large normal training set (we plan to handle such distractors in future work). 
This data set is defined by $\mathcal{D} = \{ (\mathbf{x},d,y)_i \}_{i=1}^{|\mathcal{D}|}$, where $\mathbf{x}:\Omega \rightarrow \mathbb{R}^3$ denotes a colonoscopy frame ($\Omega$ represents the frame lattice), $d \in \mathbb{N}$ represents patient identification\footnote{Note that the data set has been de-identified, so $d$ is useful only for splitting $\mathcal{D}$ into training, testing and validation sets in a patient-wise manner.}, $y \in \mathcal{Y} = \{ Normal, Abnormal \}$ denotes the normal (without polyp) and abnormal (with polyp) classes.  
The distribution of this data set is as follows: 1) Training set:  a set of 13250 normal images (without polyps), denoted by $\mathcal{D}_N \subset \mathcal{D}$, and a set containing between 10 and 80 abnormal images, denoted by $\mathcal{D}_A \subset \mathcal{D}$; 2) Validation set: 100 normal images and 100 abnormal images for model selection; and 3) Testing set: 967 images, with 217 (25\% of the set) abnormal images and 700 (75\% of the set) normal images.  The patients in the testing set do not appear in the training/validation sets and vice versa. This abnormality proportion (on the testing set) is commonly defined in other anomaly detection literature~\cite{perera2019ocgan,schlegl2019f}.
These frames were obtained with the Olympus~\textregistered 190 dual focus endoscope. 

\begin{figure*}[t]
\small
\begin{center}
 \includegraphics[width=.9\textwidth]{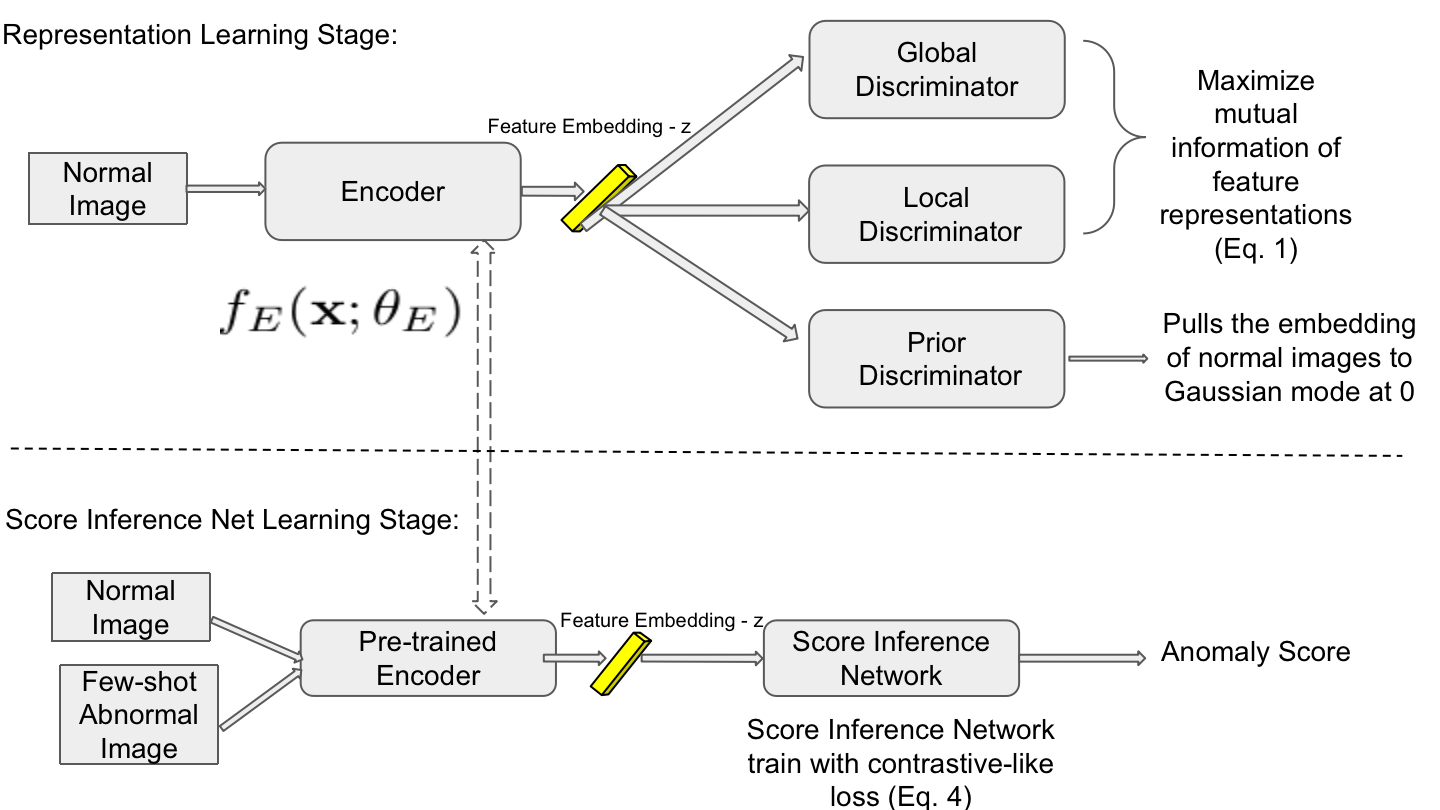}
\end{center}
\caption{The first stage of FSAD-NET training consists of modelling the encoder by maximising the MI between normal training images and embeddings in a global and local manner and by minimising the divergence of embeddings and a prior distribution~\cite{hjelm2018learning}.  The embeddings produced by the encoder are then used to train the SIN using a contrastive-like loss~\cite{pang2019deep}. }
\label{fig:structure}
\end{figure*}

\subsection{Method}
\label{sec:method_method}

The training process of our proposed FSAD-NET (Fig.~\ref{fig:auc_performance}) is divided into two stages: 
1) pre-training of a feature encoder $\mathbf{z} = f_E(\mathbf{x};\theta_E)$ ($\theta_E$ is the encoder parameter and $\mathbf{z} \in \mathbb R^Z$) to learn an image embedding that maximises the mutual information (MI) between normal images $\mathbf{x} \in \mathcal{D}_N$ and their embeddings $\mathbf{z}$~\cite{hjelm2018learning};
and 2) training of the SIN $f_S(f_E(\mathbf{x};\theta_E);\theta_S)$~\cite{pang2019deep}, parameterised by $\theta_S$, with a contrastive-like loss that uses $\mathcal{D}_N$ and $\mathcal{D}_A$ to achieve the goal 
$ f_S(f_E(\mathbf{x}\in \mathcal{D}_A; \theta_E);\theta_S) >  f_S(f_E(\mathbf{x}\in \mathcal{D}_N;\theta_E);\theta_S) $.    


More specifically, the training of the encoder to maximise the MI between the normal samples $\mathbf{x}\in \mathcal{D}_N$ and their feature embeddings $\mathbf{z} = f_E(\mathbf{x} \in \mathcal{D}_N;\theta_E)$~\cite{hjelm2018learning} is achieved with
\begin{equation}
\begin{split}
\theta_E^*,\theta_G^*,\theta_L^* = \arg\max_{\theta_E,\theta_G,\theta_L} \Big ( &
\alpha \hat{I}_{\theta_G}( \mathbf{x} ; f_E(\mathbf{x};\theta_E)) + \frac{\beta}{|\mathcal{M}|} \sum_{\omega \in \mathcal{M}} \hat{I}_{\theta_L}( \mathbf{x}(\omega) ; f_E(\mathbf{x}(\omega);\theta_E)  ) \Big ) \\
& + \gamma \arg\min_{\theta_E}\arg\max_{\phi}  \hat{D}_{\phi} (\mathbb{V} || \mathbb{U}_{\mathbb{P},\theta_E})
\end{split}
   \label{eq:train_f_E}
\end{equation}
where $\alpha,\beta,\gamma$ are the model hyperparameters, the
functions $\hat{I}_{G}(.)$ and $\hat{I}_{L}(.)$ denote an MI lower bound based on the Donsker-Varadhan representation of the Kullback-Leibler (KL)-divergence~\cite{hjelm2018learning}, defined by 
\begin{equation}
    \hat{I}_{\theta_G}(\mathbf{x};f_E(\mathbf{x};\theta_E)) = 
    \mathbb{E}_{\mathbb{J}}[f_G(\mathbf{x},f_E(\mathbf{x};\theta_E);\theta_G)]-\log \mathbb{E}_{\mathbb{M}}[e^{f_G(\mathbf{x},f_E(\mathbf{x};\theta_E);\theta_G)}],
    \label{eq:MI}
\end{equation}
with $\mathbb{J}$ denoting the joint distribution between images $\mathbf{x}$ and their respective embeddings $\mathbf{z}=f_E(\mathbf{x};\theta_E)$, $\mathbb{M}$ representing the product of the marginals of the images and embeddings, and $f_G(\mathbf{x},f_E(\mathbf{x};\theta_E);\theta_G)$ being a discriminator parameterised by $\theta_G$. 
Also in~\eqref{eq:train_f_E}, the function
$\hat{I}_{\theta_L}( \mathbf{x}(i) ; f_E(\mathbf{x}(i);\theta_E))$, defined similarly as~\eqref{eq:MI} for the discriminator $f_L(\mathbf{x}(\omega),f_E(\mathbf{x}(\omega);\theta_E);\theta_L)$,
is the local MI between image regions $\mathbf{x}(\omega)$ ($\omega \in \mathcal{M} \subset \Omega$) and respective local  embeddings $f_E(\mathbf{x}(\omega),\theta_E)$. Moreover in~\eqref{eq:train_f_E},
\begin{equation}
\arg\min_{\theta_E}\arg\max_{\phi} \hat{D}_{\phi} (\mathbb{V} || \mathbb{U}_{\mathbb{P},\theta_E}) = \mathbb{E}_{\mathbb V}[\log d(\mathbf{z};\phi)] + \mathbb{E}_{\mathbb{P}}[\log(1-d(f_E(\mathbf{x};\theta_E));\phi))],
\label{eq:div_f_E}
\end{equation}
with $\mathbb{V}$ denoting a prior distribution for the embeddings $\mathbf{z}$ ($\mathbb{V}$ is assumed to be a normal distribution $\mathcal{N}(.;\mu_{\mathbb{V}},\Sigma_{\mathbb{V}})$, with mean $\mu_{\mathbb{V}}$ and covariance $\Sigma_{\mathbb{V}}$), $\mathbb{P}$ the distribution 
of the embeddings $\mathbf{z}=f_E(\mathbf{x}\in\mathcal{N}_N;\theta_E)$, and $d(.;\phi)$ is a discriminator modelled with adversarial training to estimate the likelihood that the input is sampled from $\mathbb{V}$ or $\mathbb{P}$.  This objective function pulls the feature embeddings of the normal images toward  $\mathcal{N}(.;\mu_{\mathbb{V}},\Sigma_{\mathbb{V}})$.

The next step of the learning process consists of computing the embeddings of normal and abnormal images with $\mathbf{z} = f_E(\mathbf{x} \in \mathcal{D}_A \bigcup \mathcal{D}_N;\theta_E^*)$ to
train $f_S( \mathbf{z}; \theta_S)$ using a contrastive-like loss to directly optimise the anomaly score~\cite{pang2019deep}.  More specifically, the constrastive loss for each training sample is defined as:
\begin{equation}
    \ell_S = \mathbb{I}(y\textrm{ is }Normal)|s(f_S(\mathbf{z};\theta_S))| + \mathbb{I}(y\textrm{ is }Abnormal)\max(0,a - s(f_S(\mathbf{z};\theta_S))),
    \label{eq:loss_score_inference_net}
\end{equation}
where $\mathbb{I}(.)$ is an indicator function that is equal to one when the condition in the parameter is true, and zero otherwise, $s(x) = \frac{x - \mu_S}{\sigma_S}$ with $\mu_S=0$ and $\sigma_S=1$ representing the mean and standard deviation of the prior distribution for the anomaly scores for normal images, and $a$ is the minimum margin between $\mu_S$ and the anomaly scores of abnormal images~\cite{pang2019deep}.  The loss in~\eqref{eq:loss_score_inference_net} pulls the scores from normal images to $\mu_S$ and pushes the scores of abnormal images away from $\mu_S$ with a margin of at least $a$.

During inference, we take a test image $\mathbf{x}$, compute the feature embedding with $f_E(\mathbf{x};\theta_E)$ and then compute the score with $s = f_S(\mathbf{z};\theta_S)$ -- the score result $s$ is then compared to a threshold $\tau$ to determine if the test image is normal or abnormal. We considered the score  $s$ as the estimation of the notion of closeness which is related to the likelihood that the embedding of a colonoscopy image is classified as belonging to the set of normal images.

\section{Experiment}

\subsection{Experimental Setup}

The original colonoscopy images are resized
from initial resolution $1072 \times 1072 \times 3$ to $64 \times 64 \times 3$ to reduce the training and inference computational costs. We found that $64 \times 64 \times 3$ is the minimum size that we can use without a negative impact on AUC. We note the polyps are still visible at such resolution, as shown in Fig~\ref{fig:result_img}.
The model selection (to select optimiser, learning rate and model structure) is done using the validation set mentioned in Sec.~\ref{sec:method_data}.
We use Adam~\cite{kingma2015adam} optimiser during training with a learning rate of 0.0001 for the encoder and SIN learning. 
We adopt batch normalisation for both stages.  
We make sure our method uses a similar backbone architecture as  other competing approaches in Tab.~\ref{tab:result}. 
In particular, the encoder $f_E(.;\theta_E)$ uses four convolution layers (with 64, 128, 256, 512 filters of size 4 $\times$ 4). 
The global discriminator $f_G(.;\theta_G)$ has three convolutional layers (with 128, 64, 32 filters of size 3 $\times$ 3). 
The local discriminator $f_G(.;\theta_G)$ has three convolutional layers (with 192, 512, 512 filters of size 1 $\times$ 1). 
The prior discriminator $d(.;\phi)$ has three linear layers with 1000, 200, 1 nodes per layer).
We also use the validation set to estimate $a=6$ in~\eqref{eq:loss_score_inference_net}.
In~\eqref{eq:train_f_E}, we follow the DIM paper for setting the hyper-parameters as follows~\cite{hjelm2018learning}: $\alpha=0.5$, $\gamma=1$, $\beta=0.1$. For the prior distribution for the embeddings in~\eqref{eq:div_f_E}, we set $\mu_{\mathbb V}=\mathbf{0}$ (i.e., a $Z$-dimensional vector of zeros), and $\Sigma_{\mathbb V}$ is a $Z \times Z$ identity matrix.
To train the model, we first train the encoder, local, global and prior discriminator (representation learning stage) for 6000 epochs with a mini-batch of 64 samples. 
We then train SIN for 1000 epochs, with a batch size of 64, while fixing the parameters of encoder, local, global and prior discriminator.
We implement our method using Pytorch~\cite{paszke2017automatic}. 

The detection results are measured with the area under the receiver operating characteristic curve (AUC) on the test set~\cite{schlegl2019f,perera2019ocgan}, computed by varying the inference threshold $\tau$ for the score result $s$. 

\subsection{Anomaly Detection Results}


\begin{table}[t!]
\centering
\caption{Comparison between our proposed FSAD-NET and other state of the art zero-shot and few-shot anomaly detection methods.}
\label{tab:result}
\scalebox{0.9}{
\begin{tabular}{c|c|c}
                           & Methods                                    & AUC    \\ \hline
\multirow{7}{*}{Zero-Shot} & DAE~\cite{masci2011stacked}                & 0.6384 \\
                           & VAE~\cite{doersch2016tutorial}             & 0.6528 \\
                           & OC-GAN~\cite{perera2019ocgan}              & 0.6137 \\
                           & f-AnoGAN(ziz)~\cite{schlegl2019f}          & 0.6629 \\
                           & f-AnoGAN(izi)~\cite{schlegl2019f}          & 0.6831 \\
                           & f-AnoGAN(izif)~\cite{schlegl2019f}         & 0.6997 \\
                           & ADGAN~\cite{liu2019photoshopping}          & 0.7391 \\ \hline
\multirow{11}{*}{Few-Shot} & Densenet121~\cite{huang2017densely} (40 abnormal samples)  & 0.8231 \\
                           & cross-entropy (30 abnormal samples)        & 0.6826 \\
                           & cross-entropy (40 abnormal samples)        & 0.7115 \\
                           & Focal loss (30 abnormal samples)           & 0.7038 \\
                           & Focal loss (40 abnormal samples)           & 0.7235 \\
                           & without RL (40 abnormal samples)           & 0.6011 \\
                           & Learning to Reweight~\cite{ren2018learning} (40 abnormal samples) & 0.7862 \\
                           & AE network (30 abnormal samples)           & 0.819  \\
                           & AE network (40 abnormal samples)           & 0.835  \\
                           & FSAD-NET (30 abnormal samples)             & 0.855  \\
                           & FSAD-NET (40 abnormal samples)             & \textbf{0.9033}
\end{tabular}}
\end{table}

The test set AUC results shown in Table~\ref{tab:result} are divided into zero-shot and few-shot.  The zero-shot rows show results obtained from the following zero-shot anomaly detection methods\footnote{Codes were downloaded from the authors' Github pages and tuned for our problem.}: ADGAN~\cite{liu2019photoshopping}, OCGAN~\cite{perera2019ocgan}, f-anogan and its variants~\cite{schlegl2019f} that involve image-to-image mean square error (MSE) loss (izi), Z-to-Z MSE loss (ziz) and its hybrid version (izif). 
Our FSAD-NET model outperforms all zero-shot learning methods by a large margin, showing the importance of using a few abnormal samples for training.
For the few-shot results, we consider the cases where we have 30 and 40 abnormal training images, and we test several variants of the FSAD-NET.  
We use between 30 and 40 abnormal training images because that is the range, where we observe that our FSAD-NET produces stable AUC results.
As a baseline approach, we train Densenet121~\cite{huang2017densely} using high levels of data augmentation to deal with the training imbalance issue. However, our FSAD-Net outperforms Densenet121 by a large margin.  
The variants of FSAD-NET are designed to test the importance of each stage of our method.  The methods labelled as 'Cross entropy' and 'Focal loss' replace the contrastive loss in~\eqref{eq:loss_score_inference_net} by the cross entropy loss (commonly used in classification problems)~\cite{goodfellow2016deep} and the focal loss (robust to imbalanced learning problems)~\cite{lin2018focal}, respectively. 
FSAD-NET shows substantially better results, indicating the importance of using a more appropriate loss function for few-shot anomaly detection. To show the importance of representation learning (RL) in FSAD-Net, we tested FSAD-Net without it, which shows much lower AUC results than competing approaches.
Also, we compared our method with a few-shot learning baseline~\cite{ren2018learning}, which proposes a learning algorithm for highly imbalanced learning problems. When used to train FSAD-Net, it achieved 78.62\% of mean AUC when training with 40 abnormal training samples. Hence our model shows more accurate results than that approach. 
Furthermore, we test the importance of DIM to train the encoder in~\eqref{eq:train_f_E} by replacing it by the deep auto-encoder~\cite{masci2011stacked} (labelled as AE network) -- results show that FSAD-NET is more accurate, indicating the effectiveness of using MI and prior distribution for learning the feature embeddings in~\eqref{eq:train_f_E}.

\begin{figure}[t!]
\small
\begin{center}
 \includegraphics[width = 0.8\linewidth]{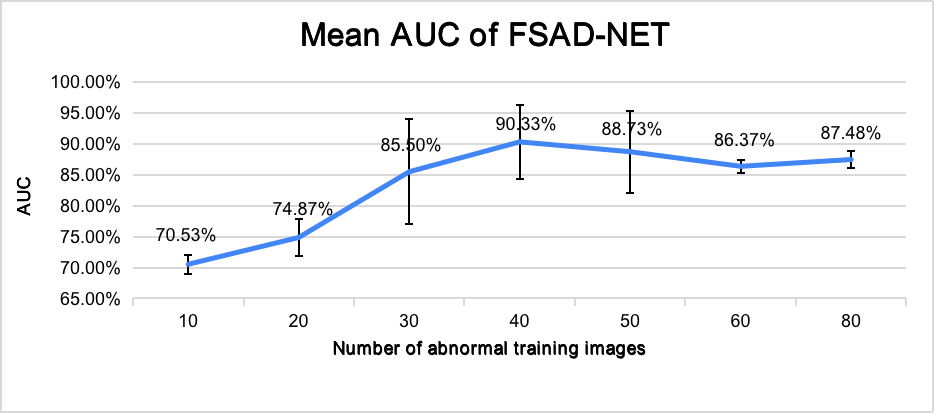}
\end{center}
\caption{AUC mean and standard deviation of FSAD-NET computed over different number of abnormal training images. 
}
\label{fig:auc_performance}
\vspace{-10pt}
\end{figure}



We further investigate the performance of our proposed FSAD-NET as a function of the number of abnormal training images that can vary from $10$ to $80$. 
For each number of abnormal training images, we train our model three times, using different training sets each time, and we compute the mean and standard deviation of the AUC results. The result of this experiment in Fig.~\ref{fig:auc_performance} shows that: 1) the performance stabilises between 85\%-90\% when feeding the model 30 or more abnormal training images; and 2) our method is robust to extremely small training sets of abnormal images.
We show a few true positive, true negative, false positive and false negative results produce by FSAD-NET in Fig.~\ref{fig:result_img}.

\begin{figure}[t!]
\small
\begin{center}
 \includegraphics[width = 0.8\linewidth]{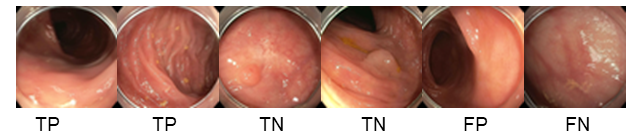}
\end{center}
\caption{True positive (TP), true negative (TN), false positive (FP) and false negative(FN) results produce by FSAD-NET (Negative = frame with polyp).}
\label{fig:result_img}
\vspace{-10pt}
\end{figure}

\section{Conclusion}

We propose the first few-shot anomaly detection framework, named as FSAD-NET, for medical image analysis applications.  FSAD-NET consists of an encoder trained to maximise the mutual information between normal images and respective embeddings and a score inference network that classifies between normal and abnormal colonoscopy frames. 
Results show that our method achieves state-of-the-art anomaly detection performance on our  colonoscopy data set, compared to previous zero-shot anomaly detection methods and imbalanced learning methods. 
In the future, we expect to extend our approach to polyp localisation and to work with colonoscopy frames containing distractors, like feces and water.

%
%
\bibliographystyle{splncs04}
%
\bibliography{bibli}

\newpage
\beginsupplement
\setcounter{section}{0}
\section{Supplementary Material}
\setcounter{equation}{0}
\setcounter{figure}{0}
\setcounter{table}{0}
\setcounter{page}{1}
\makeatletter
\renewcommand{\theequation}{S\arabic{equation}}
\renewcommand{\thefigure}{S\arabic{figure}}

\begin{figure}[]
\begin{center}
 \includegraphics[width = 1\linewidth]{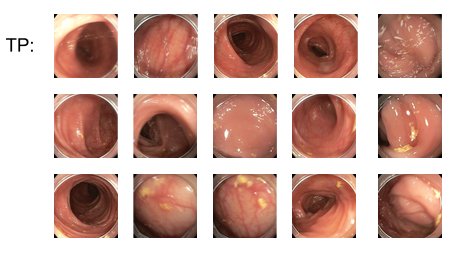}
\end{center}
\caption{True positive (TP) results produce by FSAD-NET.}
\label{fig:result_img}
\vspace{-10pt}
\end{figure}

\begin{figure}[]
\begin{center}
 \includegraphics[width = 1\linewidth]{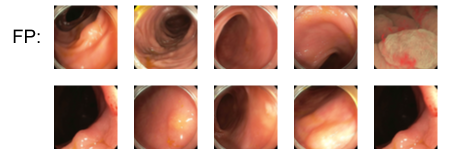}
\end{center}
\caption{ False positive (FP) results produce by FSAD-NET.}
\label{fig:result_img}
\vspace{-10pt}
\end{figure}

\begin{figure}[]
\begin{center}
 \includegraphics[width = 1\linewidth]{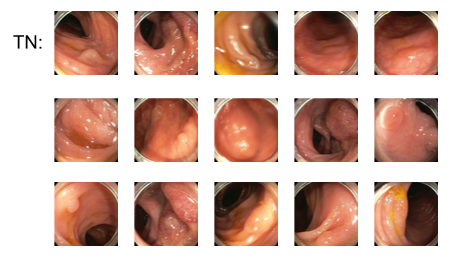}
\end{center}
\caption{True negative (TN) results produce by FSAD-NET (Negative = frame with polyp).}
\label{fig:result_img}
\vspace{-10pt}
\end{figure}

\begin{figure}[]
\begin{center}
 \includegraphics[width = 1\linewidth]{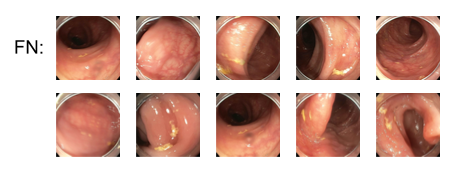}
\end{center}
\caption{ False negative(FN) results produce by FSAD-NET (Negative = frame with polyp).}
\label{fig:result_img}
\vspace{-10pt}
\end{figure}

\end{document}